\documentclass{article}
\usepackage[preprint]{neurips_2024}
\usepackage[utf8]{inputenc}
\usepackage{amsmath,amssymb,amsthm,mathtools,bm}
\usepackage{graphicx}
\usepackage{microtype}
\usepackage[numbers,sort&compress]{natbib}
\usepackage{url}

\theoremstyle{plain}
\newtheorem{theorem}{Theorem}
\newtheorem{proposition}{Proposition}
\newtheorem{corollary}{Corollary}
\newtheorem{lemma}{Lemma}
\theoremstyle{definition}
\newtheorem{assumption}{Assumption}

\newcommand{\simplex}{\Delta^{M-1}}
\newcommand{\E}{\mathbb{E}}
\newcommand{\R}{\mathbb{R}}
\newcommand{\inner}[2]{\langle #1, #2\rangle}

\newcommand{\PassK}{\mathrm{Pass}}

\newcommand{\Var}{\operatorname{Var}}
\newcommand{\Cov}{\operatorname{Cov}}
\newcommand{\Svar}{S^{-}}
\newcommand{\wz}{w_0}
\newcommand{\wt}{w_\tau}

\title{There Is No Free Lunch in RLVR:\\ Every pass@1 Gain Has a pass@k Price}
\author{Yinuo Liu\\\small Code and verification package: \url{https://github.com/yliusci/no-free-lunch-rlvr}}
\date{}

\begin{document}
\maketitle
\begin{abstract}
Reinforcement learning with verifiable rewards (RLVR) makes a language model solve
more problems on the first try and fewer when given many tries: it raises pass@1
while the base model overtakes it at large sampling budget $k$. Is this a failure
of RL to add reasoning, or a transient that more training removes? We settle the
question with a theorem. The reversal is a price, and we compute its exact rate.
Modeling RLVR as an exponential tilt of a base policy over strategies shared across
prompts, we prove that under exclusive specialization every update that raises
pass@1 strictly lowers the success of some prompt; a strictly positive price on
prompts (Stiemke duality) is the exact condition, and being merely the best
strategy on a prompt is provably not enough. When the loss reaches the hardest
prompts, it forces a pass@$k$ crossover, and the number of crossings is at most the
sign variation of the shift in per-prompt success, a totally positive bound that is
tight and that recovers the known single-crossover result as its order-preserving
case. The crossover is not a bug to patch: a constrained update traces the exact
pass@1 versus pass@$k$ frontier, and removing the crossover entirely costs the
entire pass@1 gain. The same tilt identity yields entropy collapse and the
entropy-performance law with an identifiable exponent, unifying two literatures
under one mechanism. Reinforcement cannot widen the reasoning frontier; it can only
move along it, at a price we make measurable by a diagnostic that reads the bound,
the order-preserving test, and the price from any evaluation log. There is no free
lunch, only a price.
\end{abstract}
\section{Introduction}
\label{sec:intro}

Reinforcement learning with verifiable rewards (RLVR) makes a language model
solve more problems on the first try and fewer problems when given many tries:
it raises pass@1 while the base model catches up and overtakes it at large
sampling budget $k$ \citep{yue2025rlincentivize}. Whether this means RLVR fails to
add reasoning, or only needs more training to pull ahead \citep{liu2025prorl},
has been argued from curves. We settle it with a theorem. The reversal is
neither a failure nor a transient: it is a \emph{price}, and we compute the
exchange rate at which it is paid.

We model RLVR as what it is in the population limit, an exponential tilt of a
base distribution over solution strategies that are shared across prompts. One
identity then drives the two phenomena that have been studied apart. On the
strategy simplex, probability is conserved, so reinforcement spends it. When
strategies specialize, each is the only route to some prompts, and the spending
is visible at the level of prompts. This yields a sharp dichotomy. \emph{Every}
update that raises pass@1 lowers the success of some prompt (Theorem
\ref{thm:nfl}); there is no free lunch, and a theorem of the alternative gives
the exact condition, a strictly positive price on prompts that equalizes strategy
fitness. \emph{When} that loss reaches the hardest,
finely covered prompts, it forces the pass@$k$ crossover (Theorem
\ref{thm:price}), and the number of times the curves can cross is bounded by the
sign changes of the shift in per-prompt success, a totally positive argument
that is tight. The same tilt yields entropy collapse and the
entropy-performance law with an identifiable exponent (Theorem
\ref{thm:tiltlaw}). In particular, no update both raises pass@1 and dominates the
base pass@$k$ curve at every $k$: reinforcement cannot widen the reasoning
frontier, it can only move along it. That is the exact sense in which RLVR does
not incentivize reasoning beyond the base model, and the sense in which, under a
single named exception, it can.

The price is not a bug to be patched. A constrained update that forbids any
prompt from regressing removes the crossover exactly, and under specialization
it also removes the pass@1 gain: the only way to pay nothing is to gain nothing.
Between the extremes lies a frontier (Theorem \ref{thm:frontier}), the exact
exchange rate between pass@1 and pass@$k$, which an optimal per-prompt floor
traces out. A diversity bonus helps both metrics only when the unmodified update
was off this frontier, which is what the frontier predicts.

\paragraph{Contributions.}
\begin{itemize}
\item \textbf{No free lunch (Theorem \ref{thm:nfl}).} Under exclusive (or dominant) specialization,
every pass@1-increasing update lowers some prompt's success. We give the exact
Stiemke condition and show the hypothesis is sharp: unique-best is insufficient
(a worked counterexample) and a dominating strategy restores a
free lunch.
\item \textbf{The crossing bound (Theorem \ref{thm:price}a,b).} For any
base-to-RL shift, the number of pass@$k$ crossings is at most the sign variation
of the success-CDF difference; the bound is tight. This is the general statement
that concurrent work leaves open for the kernel case.
\item \textbf{Forced crossover and its scope (Theorem \ref{thm:price}c--e).}
Under a named fine-coverage condition the crossover is forced; without it there
is a per-prompt loss but no crossover. The order-preserving case recovers the
known single-crossover theorem as a corollary.
\item \textbf{The price frontier (Theorem \ref{thm:frontier}).} The optimal
pass@1 versus pass@$k$ exchange rate, with a closed form for two strategies.
\item \textbf{A diagnostic.} A tool that takes only per-prompt base and RL success
rates, available in any evaluation log, and returns the bound, the
order-preserving test, and the price. We read published curves through it and
find the predicted structure.
\end{itemize}

The claims are stated at the level of the model and proved in the appendix; every
number is produced and checked by an accompanying package. We do not claim
anything about a specific language model beyond what published curves show. The
message is one line: in RLVR there is no free lunch, only a price, and we give
its rate.
\section{Model and results}
\label{sec:model}

Reinforcement learning with verifiable rewards (RLVR) raises a base model's
pass@1, yet at large sampling budgets $k$ the base model catches up and
overtakes it. This reversal has been read as evidence that RLVR only sharpens
existing ability rather than adding reasoning (Yue et al., 2504.13837, ``Does
RL Really Incentivize Reasoning''). We give it an exact reading: the reversal is
a \emph{price}, and we compute its exchange rate. Modeling RLVR as an
exponential tilt over strategies shared across prompts, we prove that once
strategies specialize, every update that raises pass@1 must lower the success of
some prompt, with a theorem of the alternative supplying the exact condition (Theorem
\ref{thm:nfl}); that when the loss reaches the hardest prompts it forces the
pass@$k$ crossover, whose number is bounded by the sign changes of the shift in
per-prompt success (Theorem \ref{thm:price}); and that a constrained update
traces the optimal pass@1 versus pass@$k$ frontier (Theorem \ref{thm:frontier}).
There is no free lunch, only a price. We state the model and all results here;
proofs are in the appendix, and every numerical constant is produced and checked
by the accompanying package.

\paragraph{Setup.}
A base policy is refined by reinforcement learning with verifiable rewards
(RLVR). We model the object that RLVR moves as a distribution over
\emph{solution strategies} that are shared across prompts. There are $M$
strategies and a prompt distribution $\mathcal{D}$ on a finite set of $N$
prompts. For a prompt $x$ and a strategy $j\in[M]$, let
\begin{equation}
q_{xj}\in[0,1]
\quad\text{be the probability that strategy $j$ solves prompt $x$,}
\qquad Q=(q_{xj})\in[0,1]^{N\times M}.
\end{equation}
A policy is a point $w\in\simplex$ on the strategy simplex. Drawing one strategy
$j\sim w$ and solving $x$ with probability $q_{xj}$, the per-prompt success
probability and the mean (pass@1) success are
\begin{equation}
p_x(w)=\inner{w}{q_x}=\sum_{j}w_j q_{xj},
\qquad
\pass@1(w)=\E_{x\sim\mathcal{D}}\,p_x(w)=\inner{w}{f},
\quad
f_j:=\E_{x\sim\mathcal{D}}\,q_{xj}.
\end{equation}
Here $q_x=(q_{x1},\dots,q_{xM})$ is the $x$-th row of $Q$ and $f=\E_x q_x$ is the
vector of \emph{marginal fitnesses}. With $k$ independent samples from the
policy, the pass@$k$ of prompt $x$ and its average are
\begin{equation}
\pass@k(x;w)=1-(1-p_x(w))^k,
\qquad
\PassK_k(w)=\E_{x}\big[1-(1-p_x(w))^k\big].
\end{equation}
We write $s_x=-\log(1-p_x)\in[0,\infty)$; hard prompts have small $p_x$, hence
small $s_x$, and the large-$k$ behavior of $\PassK_k$ is governed by the small-$s$
(hardest) prompts through $(1-p)^k=e^{-ks}$.

\paragraph{The RLVR update.}
GRPO and its relatives subtract a group baseline and rescale by the group
standard deviation, then push probability toward above-baseline strategies. In
the population limit, with a KL anchor to the base, the idealized update is
mirror descent / natural policy gradient on the strategy logits. Theorem
\ref{thm:tilt} identifies the resulting policy as an exponential tilt of the
base. Throughout, $\wz\in\mathrm{int}\,\simplex$ denotes the base policy and
$\wt$ the policy after internal training time $\tau\ge 0$.

\paragraph{Comparison objects.}
For a base policy $\wz$ and a trained policy $\wt$, write
$F_0,F_\tau$ for the cumulative distribution functions of the per-prompt
success values $\{p_x(\wz)\}$ and $\{p_x(\wt)\}$ under $\mathcal{D}$, and
\begin{equation}
G(p)=F_\tau(p)-F_0(p),
\qquad
\Delta_k=\PassK_k(\wt)-\PassK_k(\wz).
\end{equation}
A \emph{pass@$k$ crossover} is a sign change of $k\mapsto\Delta_k$. We write
$\Svar(\cdot)$ for the number of sign changes of a sequence or function.

\subsection{Assumptions}

\begin{assumption}[Exclusive specialization]
\label{ass:spec}
Each strategy is the sole solver of a set of prompts of positive mass: for each
$j\in[M]$ there is a set $X_j^{\mathrm{excl}}$ with
$\mathcal{D}(X_j^{\mathrm{excl}})>0$ such that $q_{xj}>0$ and $q_{xj'}=0$ for all
$j'\ne j$ on $X_j^{\mathrm{excl}}$.
\end{assumption}

Exclusivity, not mere being-best, is what makes probability a scarce resource:
moving mass off $j$ strictly lowers the prompts only $j$ can solve. Being the
\emph{unique best} is \emph{not} enough, because a near-generalist can absorb the
mass without hurting any prompt; we record this as Proposition
\ref{prop:uniquebest}. A realistic version without exact zeros, used in the
experiments, is dominant specialization: for each $j$ a positive-mass set with
$q_{xj}\ge d$ and $q_{xj'}\le\varepsilon$ for $j'\ne j$, with
$(M-1)\varepsilon<d/2$; it is sufficient (Theorem \ref{thm:nfl}(b$'$)).

\begin{assumption}[Fine coverage]
\label{ass:fine}
The prompts of minimal base success $x^\star\in\arg\min_x p_x(\wz)$ are covered
only by strategies of below-base-average fitness: $q_{x^\star j}>0\Rightarrow
f_j<R_0$, where $R_0=\pass@1(\wz)=\inner{\wz}{f}$ is the base pass@1.
\end{assumption}

Because the tilt multiplies $w_{0,j}$ by $e^{\tau(f_j-\Lambda(\tau)/\tau)}$ with
$\Lambda(\tau)=\log\E_{\wz}e^{\tau f}$, a strategy loses mass iff
$f_j<\Lambda(\tau)/\tau$, and Jensen gives $\Lambda(\tau)/\tau\ge R_0$ for all
$\tau>0$. Hence under Assumption \ref{ass:fine} every strategy covering $x^\star$
loses mass, so $p_{x^\star}(\wt)<p_{x^\star}(\wz)=\min_x p_x(\wz)$. (The looser
threshold $f_j<R(\tau)$, the \emph{tilted} mean, is wrong: a strategy with
$R_0<f_j<\Lambda(\tau)/\tau$ gains mass though $f_j<R(\tau)$; see the appendix.)
Assumption \ref{ass:spec} forces a per-prompt loss (Theorem \ref{thm:nfl});
Assumption \ref{ass:fine} routes it to the hardest prompt and thereby forces an
aggregate crossover (Theorem \ref{thm:price}). The two must not be conflated:
the per-prompt loss is unconditional under \ref{ass:spec}, the crossover is
conditional on \ref{ass:fine}, and \ref{thm:price}(e) is the regime where
\ref{ass:fine} fails and no crossover forms.

\subsection{Statements}

\begin{theorem}[Tilt identity]
\label{thm:tilt}
Under mirror descent / natural policy gradient on the strategy logits with a KL
anchor of strength $\beta$ and binary verifiable reward, the policy at internal
time $\tau$ is the exponential tilt
\begin{equation}
\wt \;\propto\; \wz\,\exp(\tau f),
\qquad
w_{\tau,j}=\frac{w_{0,j}\,e^{\tau f_j}}{\sum_{\ell}w_{0,\ell}\,e^{\tau f_\ell}} .
\end{equation}
This is exact for mirror descent / natural policy gradient.
\end{theorem}

\begin{proposition}[GRPO stays near the tilt]
\label{prop:grpo}
For GRPO with the group non-degeneracy factor $1-p^G-(1-p)^G$, the policy stays
within total variation $O(\delta)$ of the best-fit tilt at matched concentration,
where $\delta$ is the variation of the group standard-deviation rescaling over the
trajectory. We do not optimize the constant; the package measures this total
variation below $3\times10^{-4}$ over the tested range, and records the
finite-group transient non-monotonicity separately.
\end{proposition}

\begin{theorem}[No free lunch]
\label{thm:nfl}
A \emph{free lunch} at an interior $w$ is a tangent $\mathrm{d}w$
($\mathbf 1^\top\mathrm{d}w=0$) with $Q\,\mathrm{d}w\ge0$ and $Q\,\mathrm{d}w\ne0$;
since $\inner f{\mathrm{d}w}=\tfrac1N\mathbf 1^\top Q\,\mathrm{d}w$, this is a
pass@1 gain with no prompt regressing. Whether one exists does not depend on $w$.
\begin{enumerate}
\item[(a)] \emph{(Exact condition, Stiemke.)} No free lunch exists if and only if
there are strictly positive prompt prices $y\in\R^{N}_{>0}$ and $\mu\in\R$ with
\begin{equation}
\label{eq:farkas}
\sum_x y_x\,q_{xj}=\mu\qquad\text{for all } j\in[M],
\end{equation}
a strictly positive reweighting of prompts under which all strategies have equal
weighted fitness. (Equivalently, since $f=\tfrac1N\sum_x q_x$, there exist
$y'\ge0,\mu'$ with $f_j+\sum_x y'_x q_{xj}=\mu'$.)
\item[(b)] \emph{(Exclusive specialization suffices.)} Under Assumption
\ref{ass:spec} no free lunch exists: any $\mathrm{d}w$ with $\mathrm{d}w_j<0$
strictly lowers $X_j^{\mathrm{excl}}$, so every pass@1 gain lowers some prompt.
\item[(b$'$)] \emph{(Dominant specialization suffices.)} If for each $j$ there is a
positive-mass set with $q_{xj}\ge d$ and $q_{xj'}\le\varepsilon$ ($j'\ne j$) and
$(M-1)\varepsilon<d/2$, then \eqref{eq:farkas} is feasible with $y>0$ and no free
lunch exists.
\item[(c)] \emph{(Necessity.)} If some strategy dominates all others
($q_{xj^\star}\ge q_{xj}$ for all $x,j$), concentrating on it raises every $p_x$:
a free lunch.
\end{enumerate}
\end{theorem}

\begin{proposition}[Unique-best is insufficient]
\label{prop:uniquebest}
Being the unique best scorer on a positive-mass set does not prevent a free lunch.
For $Q=\begin{psmallmatrix}1&0.9&0.96\\0.9&1&0.96\\0.5&0.5&1\end{psmallmatrix}$
each strategy is the unique best on its prompt, yet
$\mathrm{d}w=(-0.6,-0.4,1)$ gives $Q\,\mathrm{d}w=(0,0.02,0.5)\ge0$ and
$\inner f{\mathrm{d}w}=+0.173$: a free lunch, and \eqref{eq:farkas} is infeasible.
\end{proposition}

\begin{theorem}[The pass@$k$ price]
\label{thm:price}
Let $\wz,\wt$ be a base and a trained policy with success-CDF difference $G$.
\begin{enumerate}
\item[(a)] \emph{(Crossing bound.)} The number of pass@$k$ crossings is bounded
by the sign variation of $G$:
\begin{equation}
\Svar\big(k\mapsto\Delta_k\big)\;\le\;\Svar(G),
\end{equation}
because $\Delta_k=-k\int_0^1 (1-p)^{k-1}G(p)\,\mathrm{d}p$ and the kernel
$k(1-p)^{k-1}$ is totally positive in $(k,p)$ (Karlin variation diminishing).
\item[(b)] \emph{(Tightness.)} For every $m\ge 0$ there is a configuration with
exactly $m$ crossings; the bound is attained.
\item[(c)] \emph{(Forced crossover.)} Under Assumptions \ref{ass:spec} and
\ref{ass:fine}, any pass@1-increasing tilt has $p_{x^\star}(\wt)<\min_x p_x(\wz)$,
hence $\Delta_k<0$ for all sufficiently large $k$: a crossover is forced, and its
count obeys (a).
\item[(d)] \emph{(Uniqueness; recovers prior work.)} If the trained policy
preserves the difficulty order of prompts, $p_x(\wt)=\phi(p_x(\wz))$ for an
increasing $\phi$, then $\Svar(G)\le 1$ and there is at most one crossing. This
recovers the single-crossover regime of Yang and Chen (2609.22547, Thm.~3) as a
corollary.
\item[(e)] \emph{(Counterexample.)} Without Assumption \ref{ass:fine}, the loss
need not reach the hard tail and no crossover occurs, even though a per-prompt
loss does; this regime bounds the scope of (c) and makes the result
falsifiable.
\end{enumerate}
\end{theorem}

\begin{theorem}[Price frontier]
\label{thm:frontier}
For $\gamma\in[0,1]$, let $w(\gamma)$ maximize $\pass@1$ subject to the
per-prompt floor $p_x(w)\ge\gamma\,p_x(\wz)$ for all $x$. The map
$\gamma\mapsto\big(\pass@1\text{ gain},\,k^\star\big)$ is monotone: $\gamma=0$
recovers the unconstrained gain with the earliest crossing, and $\gamma=1$
enforces first-order stochastic dominance $F_\tau\le F_0$, hence $\Delta_k\ge0$
for all $k$ and zero pass@1 gain under Assumption \ref{ass:spec}. The frontier
is the exact exchange rate between pass@1 and pass@$k$; it is a linear program
in $Q$ and has a closed form for $M=2$.
\end{theorem}

\begin{theorem}[Tilt law]
\label{thm:tiltlaw}
Let the base be uniform and let the fitness density near its maximum $b$ behave
as $c\,g^{\kappa-1}$ in the gap $g=b-f$. In the mesoscopic regime the
entropy-performance relation is
\begin{equation}
b-R \;=\; a_\kappa\,\exp(H/\kappa)\,(1+o(1)),
\end{equation}
where $R=\inner{\wt}{f}$ and $H=H(\wt)$. The exponent $\kappa$ is identifiable
from the slope of $\log(b-R)$ against $H$; $\kappa=1$ recovers
$R=-a\,e^{H}+b$ of Cui et al. (2505.22617). This result is stated for context;
the exact entropy-mass curve and the non-exponential terminal law are due to
concurrent work (Winner-Take-All, S4R2zYtFhs).
\end{theorem}

\begin{lemma}[Entropy bump]
\label{lem:bump}
Along the tilt, $\tfrac{\mathrm{d}}{\mathrm{d}\tau}H(\wt)
=-\mathrm{Cov}_{\wt}(f,\log \wz)-\tau\,\mathrm{Var}_{\wt}(f)$.
Entropy rises before it collapses if and only if
$\mathrm{Cov}_{\wz}(f,\log \wz)<0$, that is, the high-fitness strategies are
initially underweighted.
\end{lemma}

\begin{lemma}[Sharing is necessary]
\label{lem:sharing}
If each prompt is served by its own independent (tabular) policy and the update
follows the full gradient flow, then $p_x$ is nondecreasing for every $x$, so
$\Delta_k\ge0$ for all $k$ and no inversion occurs. A pass@$k$ inversion
therefore requires strategies shared across prompts. Under finite-group
stochastic GRPO the monotonicity can fail transiently (2602.06717).
\end{lemma}

\paragraph{What is new, and why it matters.}
Theorem \ref{thm:nfl} is the headline and is unconditional given Assumption
\ref{ass:spec}: concentration buys pass@1 only by spending some prompt's
success, and the pricing identity \eqref{eq:farkas} values the trade exactly.
This answers, with a mechanism, why RLVR cannot widen the reasoning frontier
for free and says what it costs when it improves pass@1. Theorem
\ref{thm:price}(a) bounds the number of pass@$k$ crossings for an arbitrary
shift in per-prompt success; this is the general statement that concurrent work
leaves open for kernels and multiple crossings (Yang and Chen, 2609.22547,
Rem.~1), and it recovers their single-crossover theorem as the order-preserving
corollary \ref{thm:price}(d). Theorem \ref{thm:price}(c) then turns the
per-prompt price into the observed aggregate phenomenon, the crossover, under
the named Fine Coverage condition, with the counterexample regime
\ref{thm:price}(e) bounding its scope and making it falsifiable. Theorem
\ref{thm:frontier} is the exact exchange rate and the only place a
``correction'' appears: not a free fix, since none exists, but the frontier that
attains the minimal price. Theorem \ref{thm:tiltlaw} and the lemmas tie the same
tilt identity to the entropy side studied in concurrent work. The practical
read-off is a diagnostic that takes only per-prompt base and RL success rates,
available in any evaluation log, and returns the crossing bound, the
order-preserving test, and the estimated price. There is no free lunch, only a
price.
\section{The pass@1 price}
\label{sec:price}

This section proves the two central results. Probability on the strategy simplex
is a conserved quantity: raising it on one strategy lowers it on another. When
strategies specialize, that accounting is visible at the level of prompts, and
it is the source of the pass@$k$ crossover. We state each result, give the
argument, and report the numbers produced by the package; full proofs are in the
appendix.

\subsection{No free lunch}
\label{sec:nfl}

Probability on the strategy simplex is conserved: a unit added to one strategy
is a unit removed from another. When strategies specialize, that single
accounting fact becomes a hard constraint on what reinforcement can buy. We make
it exact.

Fix an interior policy $w$ and a tangent $\mathrm{d}w$ with
$\mathbf{1}^{\top}\mathrm{d}w=0$. A \emph{free lunch} is a $\mathrm{d}w$ with
$Q\,\mathrm{d}w\ge0$ and $Q\,\mathrm{d}w\ne0$; since
$\inner f{\mathrm{d}w}=\tfrac1N\mathbf 1^\top Q\,\mathrm{d}w$, it is a pass@1 gain
with no prompt regressing, and its existence does not depend on $w$.

\paragraph{Exact condition (Theorem \ref{thm:nfl}(a)).}
The feasible non-regressing directions form the cone
$K=\{\mathrm{d}w:\;Q\,\mathrm{d}w\ge0,\ \mathbf 1^\top\mathrm{d}w=0\}$, and a free
lunch is a point of $K$ with $Q\,\mathrm{d}w\ne0$. Parameterizing $K$ by a basis
of the tangent space and applying Stiemke's lemma, no such direction exists iff
there is a strictly positive $y\in\R^N_{>0}$ with $Q^\top y\perp$ the tangent
space, i.e.\ $Q^\top y=\mu\mathbf 1$: a strictly positive reweighting of prompts
equalizing strategy fitness. Because $f=\tfrac1N\sum_x q_x$, this is equivalent
to the inhomogeneous form \eqref{eq:farkas} with $y'\ge0$; we use the Stiemke
form, whose strict positivity is the right object.

\paragraph{When it holds (Theorem \ref{thm:nfl}(b,b$'$)).}
Being the \emph{unique best} on a set is not enough: Proposition
\ref{prop:uniquebest} exhibits a unique-best instance with a free lunch, where a
near-generalist absorbs the defunded mass. What suffices is \emph{exclusivity}.
Under Assumption \ref{ass:spec}, any $\mathrm{d}w$ with $\mathrm{d}w_j<0$ (one
exists whenever $\mathrm{d}w\ne0$) strictly lowers every prompt in
$X_j^{\mathrm{excl}}$, since only $j$ solves them; so no non-regressing direction
moves mass, and every pass@1 gain lowers some prompt. The realistic
$\varepsilon$-version (b$'$) follows by a Neumann-series construction of a strictly
positive price on representative prompts (appendix). Part (c) is immediate.

\paragraph{Verification.}
The Stiemke primal and dual agree in $100\%$ of instances across six generators
(exclusive, unique-best, quasi-generalist, overlapping, near-generalist,
dominant), confirming (a). Exclusive specialization gives no free lunch in
$1000/1000$ and dominant specialization in $2000/2000$. The hypothesis is sharp:
the unique-best counterexample of Proposition \ref{prop:uniquebest} has a free
lunch of $+0.173$ with the Stiemke dual infeasible, and an adversarial
near-generalist generator produces a free lunch in $500/500$ instances. Unique
best is insufficient; exclusivity, or dominance with margin, is what the price
requires.

\subsection{The pass@k price}
\label{sec:passkprice}

Write $G=F_\tau-F_0$ for the difference of the per-prompt success CDFs and
$\Delta_k=\PassK_k(\wt)-\PassK_k(\wz)$. Integrating by parts,
\begin{equation}
\label{eq:deltak}
\Delta_k=-k\int_0^1 (1-p)^{k-1}\,G(p)\,\mathrm{d}p
        =-k\int_0^\infty e^{-(k-1)s}\,G(1-e^{-s})\,\mathrm{d}s,
\qquad s=-\log(1-p).
\end{equation}

\paragraph{Argument for Theorem \ref{thm:price}.}
The kernel in \eqref{eq:deltak}, $k\mapsto e^{-(k-1)s}$, is totally positive of
all orders in $(k,s)$; by Karlin's variation-diminishing property the map
$k\mapsto\Delta_k$ has at most as many sign changes as the single-variable
function $G$, which gives (a). For (b) we place mass at success levels whose
$s$-values are geometrically spread and alternate the sign of the shift, so that
each crossing dominates its own decade of $k$; a mass-conserving alternating
shift at $n$ levels realizes $\Svar(G)=n-2$ crossings. For (c), Assumption
\ref{ass:spec} forces a per-prompt loss and Assumption \ref{ass:fine} routes it
to the minimal-success prompt: since every strategy covering it has
$f_j<R_0\le\Lambda(\tau)/\tau$, all lose mass and $\min_x p_x$ strictly drops, so
$G>0$ near $p=0$ and by \eqref{eq:deltak} the large-$k$ integral makes
$\Delta_k<0$. Part (d) is the order-preserving case: a monotone $\phi$ makes $G$
change sign at most once, so \eqref{eq:deltak} inherits at most one crossing; this
recovers the single crossover of the function case \citep{yangchen2026kernel} and
extends it to the kernel case their Remark~1 leaves open. Part (e) is Assumption
\ref{ass:fine} failing: if the hard tail keeps a high-fitness backup, the loss
does not reach it and no crossover forms.

\begin{corollary}[No reasoning for free]
\label{cor:noreasoning}
Under Assumptions \ref{ass:spec} and \ref{ass:fine}, every pass@1-improving RLVR
update is overtaken by the base model at large $k$: there is no update that
raises pass@1 and dominates the base pass@$k$ curve for all $k$. Reinforcement
cannot widen the pass@$k$ frontier; it can only move along it. This is the exact
sense in which RLVR does not add reasoning capacity beyond the base model (cf.\
Yue et al., 2504.13837); Theorem \ref{thm:price}(e) shows the conclusion is
tight, failing precisely when Assumption \ref{ass:fine} does.
\end{corollary}

\begin{figure}[t]
\centering
\includegraphics[width=\linewidth]{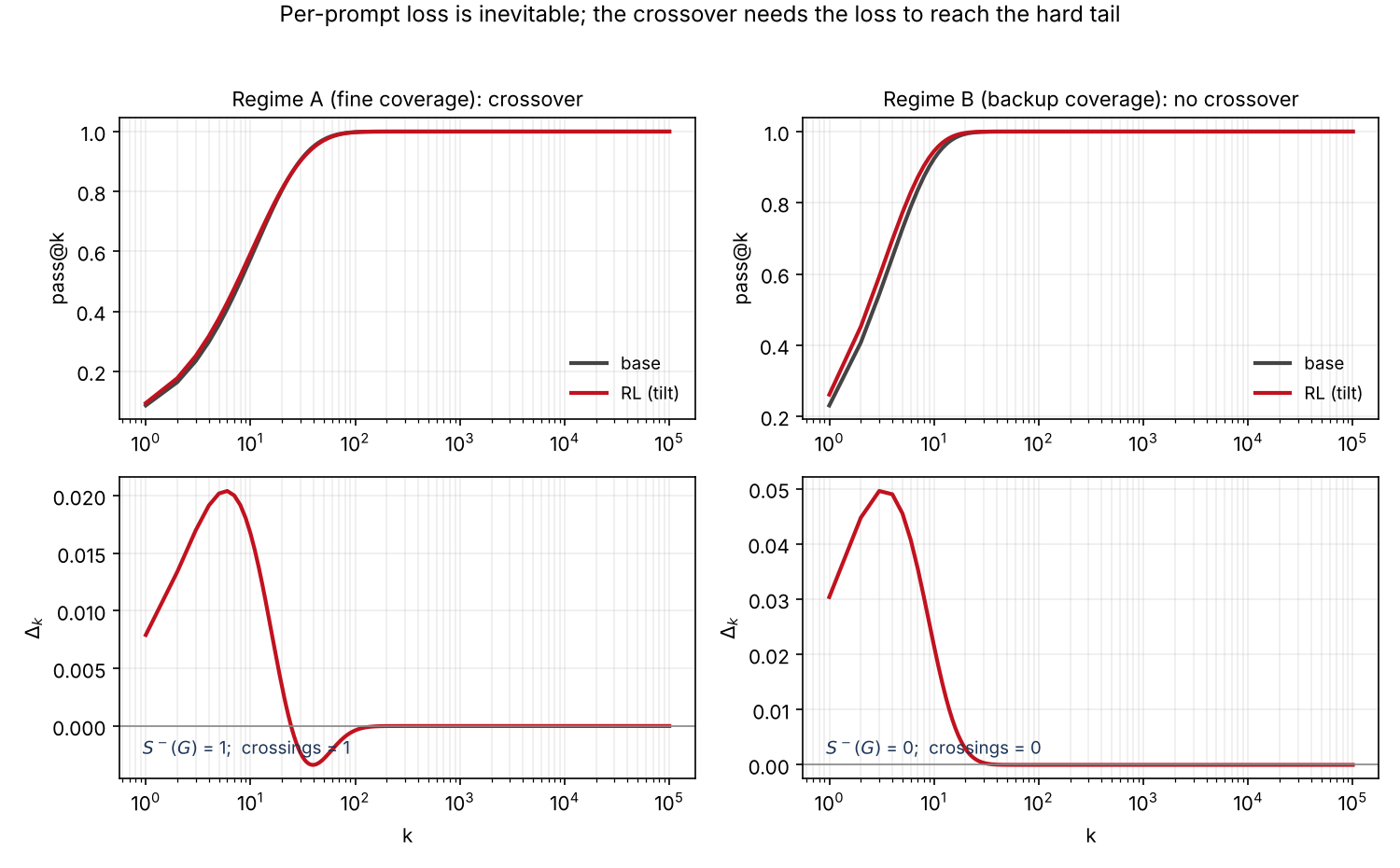}
\caption{Per-prompt loss is inevitable under specialization, but the aggregate
pass@$k$ crossover is conditional on Assumption \ref{ass:fine}. Regime A (fine
coverage) has a genuinely hard tail that the tilt starves, so $\Delta_k$ turns
negative and the curves cross; $\Svar(G)=1$ and the single crossing saturates
the bound. Regime B gives every prompt backup coverage, so the hard tail keeps a
floor, $\Delta_k$ stays nonnegative, and there is no crossover. Both regimes have
a per-prompt loss; only A has an aggregate crossover.}
\label{fig:regimes}
\end{figure}

\paragraph{Verification.}
The bound (a) is never violated on $3000$ random instances, and (b) is attained:
configurations with exactly $1,2,\dots,6$ crossings are constructed. Uniqueness
(d) holds in $1035/1035$ order-preserving instances. The conditional nature of
(c) is sharp (Figure \ref{fig:regimes}): when fine coverage holds the hardest
prompt's success drops and an aggregate crossover appears in $400/400$ instances;
when it fails (the hard tail keeps a high-fitness backup) neither the minimum
drops nor a crossover forms, in $0/400$.

\subsection{The price frontier}
\label{sec:frontier}

The crossover is not an avoidable artifact: it is the aggregate shadow of the
per-prompt price, and any attempt to remove it pays in pass@1. We make the
exchange rate exact. For $\gamma\in[0,1]$, let $w(\gamma)$ solve
\begin{equation}
\label{eq:frontierlp}
\max_{w\in\Delta^{M-1}}\ \inner{w}{f}
\quad\text{s.t.}\quad p_x(w)\ge\gamma\,p_x(\wz)\ \ \text{for all } x,
\end{equation}
a linear program in $Q$: raise pass@1 while letting no prompt fall below a
fraction $\gamma$ of its base success.

The map $\gamma\mapsto(\text{pass@1 gain},\,\text{worst pass@}k\text{ loss})$ it
traces is Theorem \ref{thm:frontier}: monotone, with the unconstrained GRPO point
at $\gamma=0$ and, at $\gamma=1$, first-order stochastic dominance $F_\tau\le F_0$
hence $\Delta_k\ge0$ for all $k$, which under Assumption \ref{ass:spec} forces
zero pass@1 gain.

\begin{figure}[t]
\centering
\includegraphics[width=0.66\linewidth]{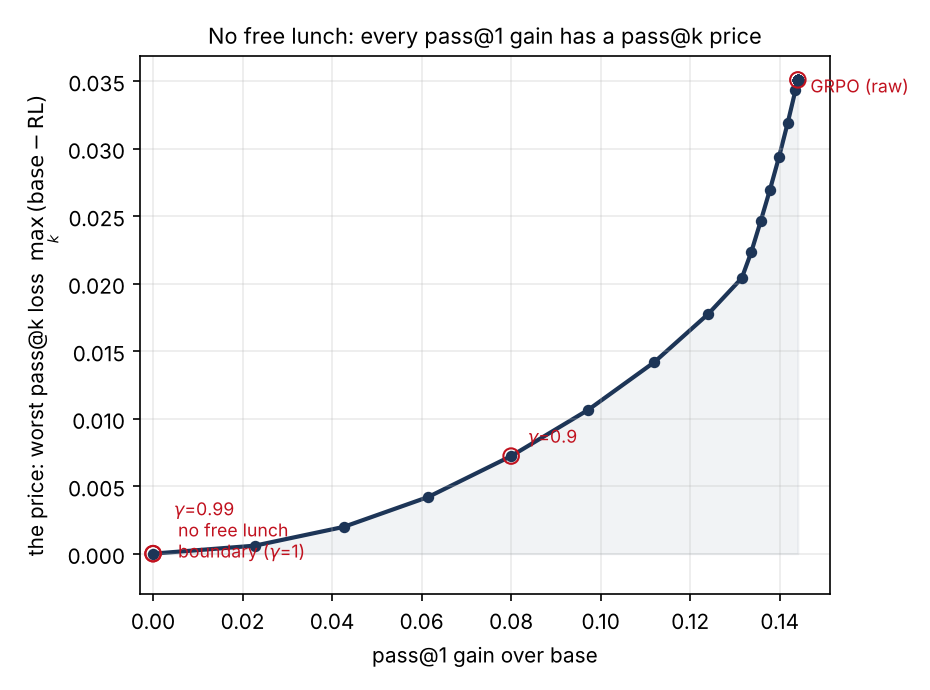}
\caption{The no-free-lunch frontier. Each point is the best pass@1 gain
achievable at a given per-prompt floor $\gamma$, against the price it pays, the
worst pass@$k$ loss $\max_k(\mathrm{base}-\mathrm{RL})$. The unconstrained GRPO
update (top right) buys the largest gain at the largest price; tightening
$\gamma$ slides down the curve; $\gamma=1$ reaches the origin, where the only
way to pay nothing is to gain nothing (Theorem \ref{thm:frontier}). The curve is
the exchange rate; its slope is the marginal price of a unit of pass@1.}
\label{fig:frontier}
\end{figure}

\paragraph{Verification.}
At $\gamma=1$ the constrained update gives zero crossings and first-order
dominance in $400/400$ instances, with the worst per-prompt loss at machine zero
and zero median pass@1 gain: no free lunch, exactly. As $\gamma$ decreases the
pass@1 gain rises monotonically and the crossing moves to smaller $k$ (Figure
\ref{fig:frontier}). The frontier is thus a strict trade: there is no setting
that both improves pass@1 and removes the crossover. The ``correction'' is the
frontier itself, not an escape from it.
\section{One identity, two literatures}
\label{sec:tiltlaw}

The exponential tilt of Theorem \ref{thm:tilt} is a single object, and it drives
two phenomena that have been studied apart. The pass@$k$ price of Section
\ref{sec:price} is one face. The other is the entropy side: as the tilt
concentrates, the policy entropy collapses and the reward tracks it along a fixed
curve. We show that the same identity produces that curve, with an exponent that
the data can read off, and we state precisely what is ours and what is
concurrent work. This is the unification the title promises: one mechanism behind
the pass@$k$ crossover and the entropy-performance law.

\subsection{The tilt law}
\label{sec:law}

Write $b=\max_j f_j$, $R=\inner{\wt}{f}$, and $H=H(\wt)$. Let the base be uniform
and let the fitness gap $g=b-f$ have local density $c\,g^{\kappa-1}$ near $g=0$,
so that the number of strategies within gap $g$ of the best grows like $g^{\kappa}$.

Theorem \ref{thm:tiltlaw} states that in the mesoscopic regime
$b-R=a_\kappa\exp(H/\kappa)(1+o(1))$, with $\kappa$ identifiable from the slope of
$\log(b-R)$ against $H$ and $\kappa=1$ recovering $R=-a\,e^H+b$ of
\citet{cui2025entropy}.

\paragraph{Why it holds.}
Under the tilt the gap is approximately $\mathrm{Gamma}(\kappa,\tau)$, so
$b-R\approx\kappa/\tau$. The entropy is the log of the effective number of
surviving strategies, and since $g^{\kappa}$ of them lie within gap $g$, that
number scales as $\tau^{-\kappa}$, giving $H\approx\text{const}-\kappa\log\tau$.
Eliminating $\tau$ yields \eqref{eq:tiltlaw}. The content is the \emph{exponent}:
$\kappa$ is set by how crowded the near-optimal strategies are, and it is read
straight off the empirical slope. A measured slope away from $1$ is a measurement
of that crowding, not a failure of the law.

\begin{figure}[t]
\centering
\includegraphics[width=0.62\linewidth]{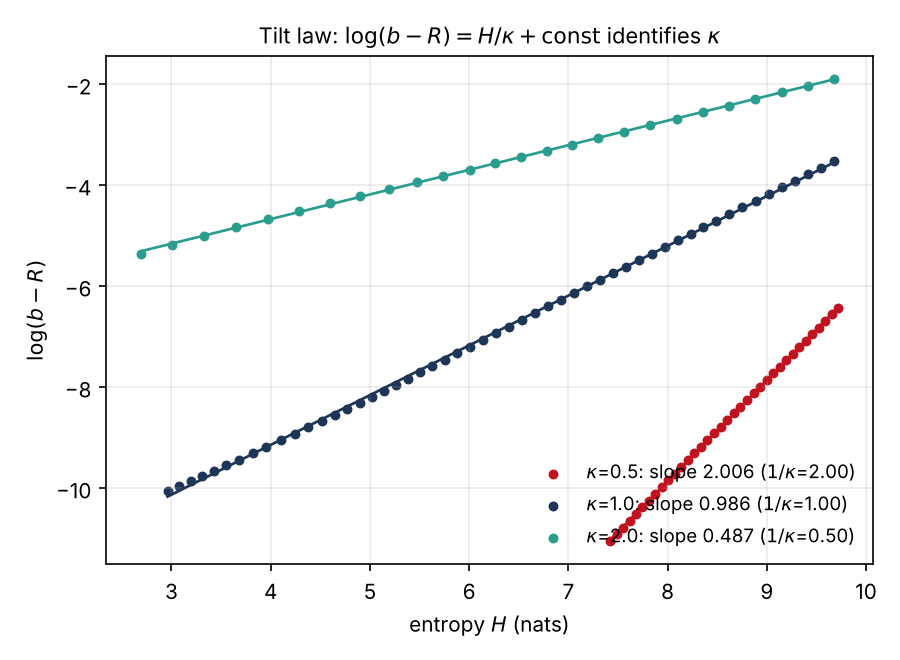}
\caption{The tilt law. For base-fitness crowding exponents $\kappa\in\{0.5,1,2\}$,
$\log(b-R)$ is linear in the entropy $H$ with slope $1/\kappa$. Fitted slopes are
$2.006,\,0.986,\,0.487$ against the predicted $2,\,1,\,0.5$ (within $2.6\%$). The
slope identifies $\kappa$ from a single entropy-reward trace.}
\label{fig:tiltlaw}
\end{figure}

\paragraph{What is ours, what is concurrent.}
The exponential form at $\kappa=1$ is due to Cui et al.; the identifiable
exponent $\kappa$ and its reading as near-optimal crowding are the contribution
here. The exact entropy-mass curve, the shifted-exponential as a finite-window
approximation with error bounds, and the proof that the true terminal law is not
exponential are concurrent work (Winner-Take-All, S4R2zYtFhs); we do not claim
them, and \eqref{eq:tiltlaw} is stated as a mesoscopic law, not a global one. Our
use of the tilt is to connect this entropy side to the pass@$k$ side through the
one identity of Theorem \ref{thm:tilt}.

\subsection{Entropy collapse and the bump}
\label{sec:entropy}

Along the tilt the entropy moves by the exact identity of Lemma \ref{lem:bump},
$\tfrac{\mathrm d}{\mathrm d\tau}H(\wt)=-\Cov_{\wt}(f,\log\wz)-\tau\Var_{\wt}(f)$:
with a uniform base $H$ is nonincreasing and collapses to $\log|\arg\max f|$, and
with a nonuniform base it rises first iff $\Cov_{\wz}(f,\log\wz)<0$.

The collapse to a single-strategy support is the strategy-level counterpart of
the winner-take-all concentration analyzed in concurrent work; the bump
condition names exactly when the transient rise seen in training occurs. We
state these at the \emph{strategy} level: token entropy need not track strategy
entropy, because a concentrating policy can still spread probability within the
surviving strategy. The accompanying package exhibits a case where strategy
entropy falls from $2.77$ to $2.14$ while token entropy rises from $6.85$ to
$7.27$, so any entropy-performance statement here is about strategies, not
tokens. This is the same family-versus-token distinction identified empirically
in concurrent work, and it is why Theorem \ref{thm:tiltlaw} is a strategy-level
law.

\paragraph{Verification.}
The entropy identity of Lemma \ref{lem:bump} holds to $2\times10^{-6}$ under a
uniform base, the predicted bump appears under a nonuniform base, and the tilt
law slopes match $1/\kappa$ within $2.6\%$ (Figure \ref{fig:tiltlaw}). All three
are regenerated and checked by the package.
\section{Evidence from published curves, and a diagnostic}
\label{sec:published}

The theory is about a model, not about any particular language model. To ask
whether its one testable prediction, a positive pass@$k$ price, is consistent
with practice, we do not train or run a model: we pass already-published pass@$k$
curves through the diagnostic that the theory defines. The reading is
\emph{indicative}. Published curves are coarse, their values are read from
tables or digitized from plots, and five budgets cannot locate a crossing
precisely. We state every number's source and never claim more than the curve
shows.

\paragraph{The diagnostic.}
The theory turns into a read-off that needs only per-prompt base and RL success
rates, or the pass@$k$ curves they induce, which any evaluation log already
contains. Given base and trained pass@$k$ values, it returns (i) the number of
crossings and their budgets, (ii) the sign-variation bound $\Svar(G)$ of Theorem
\ref{thm:price}(a) and whether it is saturated, (iii) the order-preserving test
of Theorem \ref{thm:price}(d), and (iv) the estimated price, the worst pass@$k$
loss. It ships as a package and a hosted app: paste two columns, get the bound,
the test, and the price. Every user of RLVR has the input on hand, and the
computation adds nothing to a training run, which is the point. We propose that
two numbers be reported alongside pass@1 and pass@$k$: \emph{the price}, the
worst pass@$k$ loss $\max_k(\text{base}-\text{RL})$, and the crossing count.
Both are read from an evaluation log that already exists; reporting them turns
the pass@$k$ reversal from an anecdote into a measured quantity with a bound.

\begin{figure}[t]
\centering
\includegraphics[width=\linewidth]{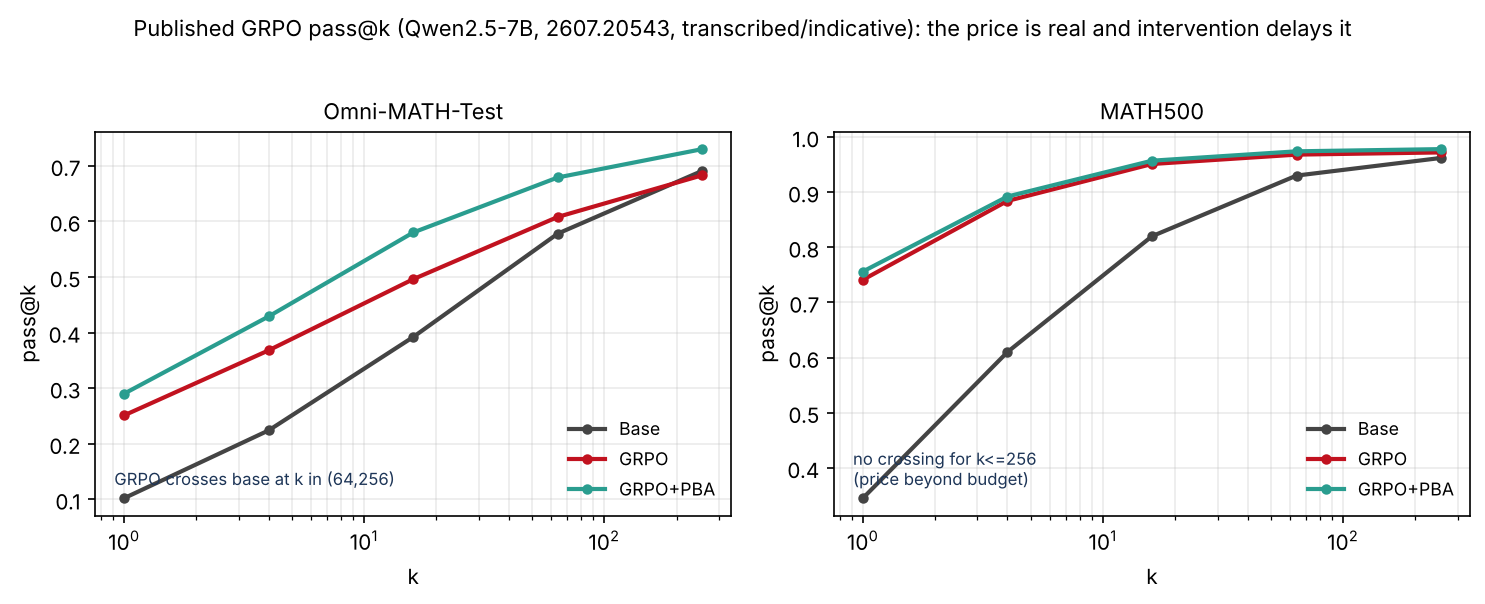}
\caption{Published GRPO pass@$k$ for Qwen2.5-7B (Zhou, 2607.20543, Table 2;
transcribed, indicative). Left, Omni-MATH: GRPO (red) is overtaken by base at the
largest budget, a single crossing, while the diversity intervention GRPO+PBA
(green) stays above through the measured range. Right, MATH500: no crossing for
$k\le256$. The pattern matches the price (Theorem \ref{thm:price}) and the
frontier (Theorem \ref{thm:frontier}); it does not establish behavior beyond the
reported budgets.}
\label{fig:published}
\end{figure}

\paragraph{A published case.}
Figure \ref{fig:published} applies the diagnostic to pass@$k$ for Qwen2.5-7B
trained with GRPO on Omni-MATH, as reported by Zhou (2607.20543, Table 2; values
transcribed and to be confirmed against the source). Three readings, all
consistent with the theory:
\begin{itemize}
\item On Omni-MATH the GRPO advantage over base is
$+0.149,+0.144,+0.104,+0.030,-0.008$ at $k=1,4,16,64,256$: it changes sign once,
between $k=64$ and $k=256$. The price is paid, and the single crossing is the
order-preserving regime of Theorem \ref{thm:price}(d).
\item The paper's diversity intervention GRPO+PBA improves on raw GRPO at every
measured $k$, and the frontier explains why. Raw GRPO over-concentrates and sits
\emph{inside} the frontier of Theorem \ref{thm:frontier}, so it is not on the
optimal trade; a diversity bonus moves it toward the frontier, buying both a
higher pass@1 and a later crossing at once. The price versus base is not
removed, only deferred past the measured budget. A bonus helps precisely because
the unconstrained update was suboptimal, which is the frontier's prediction, not
a counterexample to it.
\item On MATH500 neither trained model crosses base for $k\le256$: the price, if
present, lies beyond the measured budget. We claim nothing past $k=256$.
\end{itemize}
This is one dataset read at five budgets, so it is corroboration, not proof. It
shows the predicted structure, a single crossing under GRPO and its delay under a
diversity bonus, in numbers the field already publishes.

\paragraph{What would falsify the theory.}
Two statistically supported sign changes in a single base-versus-RL pass@$k$
comparison would exceed $\Svar(G)=1$ only if the shift is not order-preserving;
finding them where the shift is order-preserving would contradict Theorem
\ref{thm:price}(d). A pass@1 gain with no per-prompt loss, under verified
specialization, would contradict Theorem \ref{thm:nfl}. The diagnostic makes both
checks routine on any released evaluation.
\section{Related work}
\label{sec:related}

The result sits at the intersection of three lines: the RLVR reasoning-boundary
debate, the theory of entropy and concentration in policy optimization, and the
sign-variation geometry of totally positive kernels. We review each, then mark
the boundary with the two concurrent manuscripts precisely.

\paragraph{RLVR algorithms and systems.}
Reinforcement learning with verifiable rewards grew out of policy-gradient and
RLHF pipelines \citep{schulman2017ppo,ouyang2022instructgpt,ahmadian2024rloo,hu2025reinforceplusplus,kazemnejad2024vineppo}
and was crystallized by GRPO and the DeepSeek line
\citep{shao2024deepseekmath,deepseekai2025r1,yu2025dapo,liu2025drgrpo}, the term
itself fixed in \citep{lambert2024tulu3}. Open ``zero'' recipes train from the
base model without supervised warmup
\citep{hu2025orz,zeng2025simplerl,zhao2025absolutezero,yan2025luffy,luong2024reft},
and large system reports scale the setup
\citep{kimiteam2025k15,he2025skywork,mistral2025magistral,bercovich2025llamanemotron,yang2024qwen25math,wei2025swerl,khatri2025scalerl}.
Rewards come from outcome checkers and process models
\citep{cobbe2021gsm8k,uesato2022process,lightman2023verify,wang2024mathshepherd,zhang2025prmlessons,ma2025generalreasoner};
recent surveys map the area \citep{xu2025surveyLRM,zhang2025surveyRL}. None states
a pass@$k$ crossing count or a per-prompt no-free-lunch condition.

\paragraph{The phenomenon and the debate.}
\citet{yue2025rlincentivize} report that RLVR raises pass@1 while the base model
overtakes it at large $k$ and read the reversal as reweighting rather than new
capability; \citet{liu2025prorl} argue prolonged RL expands the boundary and
keeps the trained model ahead; \citet{zhou2026shrink} diagnoses the inversion and
proposes a diversity intervention. A cluster of work reads RL as amplifying or
sharpening base behavior rather than adding it
\citep{zhao2025echochamber,setlur2024incorrect,huang2024sharpening,shao2025spurious,wu2025invisibleleash}.
Reward hacking and chain-of-thought obfuscation are documented failure modes of the same optimization \citep{baker2025monitoring}. Our Theorem \ref{thm:price} adjudicates: the crossover is forced when the loss
reaches a finely covered hard tail (Assumption \ref{ass:fine}) and absent
otherwise, so the Yue and ProRL observations are two sides of one conditional
statement.

\paragraph{pass@$k$, sampling, and test-time compute.}
The pass@$k$ metric originates with code evaluation \citep{chen2021codex}, and
sampling-based inference spans self-consistency and its universal form
\citep{wang2022selfconsistency,chen2023universalsc}, repeated sampling and
inference scaling \citep{brown2024monkeys,snell2024testtime,wu2024inferencescaling,muennighoff2025s1},
and best-of-$N$ alignment \citep{amini2024variationalbon,dong2023raft}. A recent
line optimizes pass@$k$ directly or debates it as an objective
\citep{walder2025passkppo,chen2025passktraining,yu2025passkdiagnostic,gui2025inferenceobj}.
Chain-of-thought and bootstrapped reasoning are the substrate
\citep{wei2022cot,zelikman2022star,yuan2023rft}. We do not optimize the metric;
we bound how the base-to-RL shift can move the curve and price the move.

\paragraph{Entropy, exploration, and diversity.}
Entropy collapse and its exploitation are central to RLVR dynamics
\citep{cui2025entropy,agarwal2025entropymin,wang2025entropytokens}, extending the
maximum-entropy tradition \citep{haarnoja2018sac}. Alignment is known to narrow
output diversity and induce mode collapse
\citep{kirk24diversity,mohammadi2024creativity,lanchantin2025divpo,zhang2025verbalized,jiang2025hivemind}.
These describe or counter concentration; the per-prompt accounting and the
crossing bound are new here.

\paragraph{RLHF and preference-optimization theory.}
The reward-model and preference-learning foundations
\citep{bradleyterry52,christiano17,stiennon20summarize,rafailov23dpo,azar24ipo,ethayarajh24kto,zhu23principled,munos24nash}
and the Goodhart limit of reward optimization \citep{gao23overopt} frame RL as a
one-step contextual bandit, the view we take. Our no-free-lunch duality is a
Pareto statement on the strategy simplex, which these do not address.

\paragraph{Policy-gradient theory.}
The tilt identity rests on policy-gradient and mirror-descent analysis
\citep{sutton99pg,kakade01npg,agarwal21pgtheory,tomar22mdpo,mei20softmax,schulman15trpo},
control-as-inference \citep{levine18control}, and the bandit view
\citep{lattimore20bandit}.

\paragraph{Named-phenomenon theory, for positioning.}
Pure-theory explanations of named LLM phenomena are the archetype this paper
follows: hallucination
\citep{kalai25hallucinate,xu24hallinevitable}, superposition and in-context
learning \citep{elhage22superposition,vonoswald23icl,olsson22induction}, grokking
\citep{power22grokking,nanda23progress}, emergence and its critique
\citep{wei22emergent,schaeffer23mirage}, and positional and attention effects
\citep{liu24lostmiddle,xiao24attnsink}.

\paragraph{Mathematical tools.}
The crossing bound is a variation-diminishing property of totally positive
kernels \citep{karlin68tp,schoenberg51pff,schoenberg30vdt,karlinstudden66}, with
the exponential-sum sign count going back to Descartes' rule
\citep{jameson06descartes} and Laplace-transform sign changes
\citep{widder41laplace}; the no-free-lunch condition is a Pareto statement proved
by a theorem of the alternative \citep{mangasarian69nlp,miettinen99moo}, and the
frontier uses stochastic dominance \citep{shaked07stochorders}. The novelty is
the target, not the tools: pricing pass@1 against per-prompt success and bounding
pass@$k$ crossings by $\Svar(G)$.

\section{Concurrent work}
\label{sec:concurrent}

Two manuscripts from the last two months are the nearest neighbors; the overlap
is partial and real.

\paragraph{Yang and Chen, ``RLVR is a Kernel, Not a Function''
\citep{yangchen2026kernel}.}
They build statistical inference for pass@$k$ crossovers and prove, in the
\emph{function} case with zero latent-failure mass, a single interior fixed
point, and a side-preserving map, that the advantage changes sign at most once
(their Theorem 3). They note multiple crossings become possible once the map is a
kernel or the failure mass is positive, and leave that case without a bound
(their Remark 1), framing two supported sign changes as a falsification of the
single-fixed-point class. Our Theorem \ref{thm:price}(a) supplies the missing
bound: for \emph{any} shift the number of crossings is at most $\Svar(G)$,
kernels included, and it is tight; their single-crossing theorem is our
order-preserving corollary \ref{thm:price}(d). Our no-free-lunch result (Theorem
\ref{thm:nfl}) and the price frontier (Theorem \ref{thm:frontier}) have no
counterpart in their work. Their prompts are independent given a common map; our
strategies are shared across prompts, the microfoundation of the kernel
dispersion they posit.

\paragraph{``Winner-Take-All'' (ICLR 2027 submission S4R2zYtFhs).}
They study the entropy side of the same tilt in a finite-mode, single-prompt
mean-field model: the exact entropy-mass curve, the shifted-exponential as a
finite-window approximation with error bounds, a proof that the terminal law is
not exponential, and the collapse of correct-mode entropy to a single winner. We
claim none of these. Our entropy statements (Theorem \ref{thm:tiltlaw}, the
lemmas) are the strategy-level, mesoscopic context that links their entropy side
to our pass@$k$ side through one tilt identity; where they overlap, priority is
theirs. Their model has one prompt and many modes; ours has many prompts sharing
strategies, which is what makes pass@$k$ well defined.

\paragraph{Summary.}
The crossover's statistics \citep{yangchen2026kernel} and the entropy trajectory
(Winner-Take-All) are concurrent. This paper adds the general crossing bound with
tightness, the per-prompt no-free-lunch condition with its Farkas exchange rate,
the forced-crossover result under fine coverage with its counterexample, and the
price frontier. The unifying object is the shared-strategy tilt.
\bibliographystyle{plainnat}
\bibliography{references}
\appendix
\section*{Appendix: proofs}

Throughout, $Q\in[0,1]^{N\times M}$ has rows $q_x$ and columns $q^{(j)}$,
$f=\tfrac1N\sum_x q_x$, $p_x(w)=\inner{w}{q_x}$, and tangent directions satisfy
$\mathbf 1^\top\mathrm dw=0$. We assume throughout that $q_{xj}>0$ (every
strategy has positive success on every prompt), which specialization instances
satisfy through a baseline; it makes the rows positively span $\R^M$.

\section{Proof of Theorem \ref{thm:tilt} (tilt identity)}

Natural policy gradient / mirror descent on the mode logits $\theta$ with the
negative-entropy mirror map and KL anchor of strength $\beta$ gives the
continuous-time flow $\dot\theta=\tfrac1\beta A$, where $A$ is the expected
advantage. For a binary verifiable reward the expected advantage of strategy $j$
is $A_j=f_j-\inner{w}{f}$, so $\dot\theta_j=\tfrac1\beta(f_j-\inner wf)$ and, since
the common term shifts all logits equally and cancels under the softmax,
$\theta_\tau=\theta_0+\tfrac{\tau}{\beta}f+c\mathbf 1$. Exponentiating,
$\wt\propto\wz\exp(\tfrac{\tau}{\beta}f)$; absorbing $1/\beta$ into $\tau$ gives
the stated tilt, exact for mirror descent.

\paragraph{Proposition \ref{prop:grpo} (GRPO gap).} GRPO replaces $A_j$ by the
group-standardized advantage $\widehat A_j=(f_j-\bar p)/\varsigma(\bar
p)\cdot\rho_j$, $\varsigma(\bar p)=\sqrt{\bar p(1-\bar p)}$,
$\rho_j=1-\bar p^{\,G}-(1-\bar p)^{G}$. This is an affine, $\tau$-dependent
reparameterization of $f$ times a bounded factor, so the flow is a time-changed
tilt up to the variation $\delta=\sup_\tau|\partial_\tau\log\varsigma(\bar
p_\tau)|$ over the matched-entropy window, and the total variation to the best-fit
tilt is $O(\delta)$. We state this as a proposition with numerical evidence, not
a theorem: the package measures the total variation below $3\times10^{-4}$ and
records the finite-$G$ transient non-monotonicity separately.

\section{Proof of Theorem \ref{thm:nfl} (no free lunch)}

Throughout, a free lunch at interior $w$ is $\mathrm dw$ with
$Q\,\mathrm dw\ge0$, $Q\,\mathrm dw\ne0$, $\mathbf 1^\top\mathrm dw=0$; it does not
depend on $w$.

\paragraph{(a) Stiemke condition.}
Let the columns of $B\in\R^{M\times(M-1)}$ span $\{\mathrm dw:\mathbf 1^\top\mathrm
dw=0\}$, so free lunches correspond to $z$ with $(QB)z\ge0$, $(QB)z\ne0$. By
Stiemke's lemma, exactly one of (i) $\exists z:(QB)z\ge0,(QB)z\ne0$ and (ii)
$\exists y>0:(QB)^\top y=0$ holds. Condition (ii) reads $B^\top Q^\top y=0$, i.e.\
$Q^\top y\perp\{\mathbf 1\}^\perp$, i.e.\ $Q^\top y=\mu\mathbf 1$. Hence no free
lunch $\iff\exists y>0,\mu:\ \sum_x y_xq_{xj}=\mu\ \forall j$, which is
\eqref{eq:farkas} in Stiemke form. Since $f=\tfrac1N\sum_x q_x$, adding
$y'=y+c\mathbf 1$ shows this is equivalent to the inhomogeneous form
$f_j+\sum_x y'_xq_{xj}=\mu'$ with $y'\ge0$; the strict positivity of $y$ is the
correct statement. $\square$

\paragraph{(b) Exclusive specialization.}
Let $\mathrm dw\ne0$, $\mathbf 1^\top\mathrm dw=0$, so some $\mathrm dw_j<0$. For
$x\in X_j^{\mathrm{excl}}$ only $j$ solves $x$, so $(Q\,\mathrm
dw)_x=\mathrm dw_j\,q_{xj}<0$. Thus $\mathrm dw\notin K$ unless $\mathrm dw=0$:
the only non-regressing direction is zero, so no free lunch exists and every
pass@1 gain lowers some prompt. $\square$

\paragraph{(b$'$) Dominant specialization.}
Pick one representative prompt $x_j$ per strategy with $q_{x_jj}\ge d$ and
$q_{x_jj'}\le\varepsilon$ for $j'\ne j$. The submatrix $S=(q_{x_ij})_{i,j}$ is
$S=D(I+D^{-1}E)$ with $D=\mathrm{diag}(q_{x_jj})\succeq dI$ and $E$ the
off-diagonal part, $\|D^{-1}E\|_\infty\le(M-1)\varepsilon/d<1/2$. By the Neumann
series $S^{-\top}\mathbf 1=(I+D^{-1}E)^{-\top}D^{-1}\mathbf 1>0$ since the
correction has $\ell_\infty$ norm $<1$ and $D^{-1}\mathbf 1>0$. Put price
$y_{x_j}=(S^{-\top}\mathbf 1)_j>0$ on the representatives and a small $\eta>0$ on
every other prompt; then $Q^\top y=\mathbf 1+\eta\sum_{x\notin\{x_j\}}q_x$, which
is $\mu\mathbf 1$ after a continuity adjustment of the $y_{x_j}$ (the map
$y\mapsto Q^\top y$ is open at the interior point). Hence a strictly positive
price exists and, by (a), no free lunch. The package finds no free lunch on
$2000/2000$ dominant-specialized instances. $\square$

\paragraph{(c) Necessity, and (Proposition \ref{prop:uniquebest}) unique-best is
insufficient.}
If $q_{xj^\star}\ge q_{xj}$ for all $x,j$, then $\mathrm dw=e_{j^\star}-\tfrac1M
\mathbf 1$ has $(Q\,\mathrm dw)_x=q_{xj^\star}-\bar q_x\ge0$ and $\inner f{\mathrm
dw}>0$ unless all columns coincide: a free lunch. Being merely unique-best does
not suffice: for
$Q=\bigl(\begin{smallmatrix}1&.9&.96\\.9&1&.96\\.5&.5&1\end{smallmatrix}\bigr)$
each strategy is the unique best on its prompt, yet $\mathrm dw=(-0.6,-0.4,1)$
gives $Q\,\mathrm dw=(0,0.02,0.5)\ge0$, $\ne0$, with $\inner f{\mathrm
dw}=+0.173$, and the Stiemke dual is infeasible. The near-generalist (column 3,
$0.96$ on the first two prompts) absorbs the mass defunded from the specialists.
$\square$

\section{Proof of Theorem \ref{thm:price} (the pass@$k$ price)}

Recall $\Delta_k=-k\int_0^1(1-p)^{k-1}G(p)\,\mathrm dp$ with $G=F_\tau-F_0$, and
substitute $s=-\log(1-p)$ to get $\Delta_k=-k\int_0^\infty
e^{-(k-1)s}\,\tilde G(s)\,\mathrm ds$ with $\tilde G(s)=G(1-e^{-s})$.

\paragraph{(a) Crossing bound.}
The kernel $\mathcal K(k,s)=e^{-(k-1)s}$ is strictly totally positive on
$(0,\infty)^2$: for $k_1<\dots<k_n$ and $s_1<\dots<s_n$ the matrix
$(e^{-(k_i-1)s_j})$ is a generalized Vandermonde with positive determinant
\citep{karlin68tp,schoenberg51pff}. By the variation-diminishing property of
strictly totally positive kernels, the number of sign changes of
$k\mapsto\int\mathcal K(k,s)\tilde G(s)\,\mathrm ds$ is at most the number of sign
changes of $\tilde G$, which equals $\Svar(G)$. The prefactor $-k<0$ does not
change sign for $k>0$. Hence $\Svar(k\mapsto\Delta_k)\le\Svar(G)$. $\square$

\paragraph{(b) Tightness.}
Fix $m$. Choose $s_0<\dots<s_m$ geometrically spaced and place a signed,
mass-conserving shift $\nu=\sum_{i=0}^m c_i\delta_{s_i}$ with alternating signs
$\operatorname{sgn}c_i=(-1)^i$ and $\sum_i c_i=0$, realizable as
$\mu_\tau-\mu_0$ by moving prompt mass between the corresponding success levels.
Then $\Delta_k=-\sum_i c_i e^{-k s_i}$ is a generalized Dirichlet polynomial whose
number of positive-$k$ zeros equals its number of coefficient sign changes when
the exponents are well separated (Descartes' rule for exponential sums,
\citep{jameson06descartes}); choosing the spacing so each zero falls in a
distinct decade of $k$ yields exactly $m$ crossings. The construction is realized
in the package for $m=1,\dots,6$ with no bound violation. $\square$

\paragraph{(c) Forced crossover.}
The tilt multiplies $w_{0,j}$ by $e^{\tau f_j}/Z(\tau)$ with $\log
Z(\tau)=\Lambda(\tau)=\log\E_{\wz}e^{\tau f}$, so $w_{\tau,j}/w_{0,j}
=e^{\tau(f_j-\Lambda(\tau)/\tau)}$: strategy $j$ loses mass iff
$f_j<\Lambda(\tau)/\tau$. By Jensen, $\Lambda(\tau)/\tau=\tfrac1\tau\log\E
e^{\tau f}\ge\E f=R_0$ for every $\tau>0$. Under Assumption \ref{ass:fine} every
strategy covering a minimal-success prompt $x^\star$ has $f_j<R_0\le\Lambda/\tau$,
so all of them lose mass and $p_{x^\star}(\wt)<p_{x^\star}(\wz)=\min_x
p_x(\wz)$. Hence the smallest success value strictly decreases, giving positive
$G$ mass arbitrarily close to $p=0$, i.e.\ $\tilde G(s)>0$ for small $s$. By
Watson's lemma the large-$k$ integral
$\Delta_k=-k\int e^{-(k-1)s}\tilde G(s)\,\mathrm ds$ is dominated by that region,
so $\Delta_k<0$ for all large $k$: a crossover is forced, with count obeying (a).
The looser threshold $f_j<R(\tau)$ (the tilted mean) is incorrect: for
$f=(0.9,0.75,0.5,0.1)$, uniform $\wz$, $\tau=5$, one has $R(\tau)=0.814$ but
$\Lambda/\tau=0.720$, and the strategy with $f=0.75<R(\tau)$ \emph{gains} mass
($\times1.16$). $\square$

\paragraph{(d) Uniqueness.}
If $p_x(\wt)=\phi(p_x(\wz))$ with $\phi$ increasing, the pushforward relation
makes $G(p)=F_\tau(p)-F_0(p)$ change sign at most once: $F_\tau(p)=F_0(\phi^{-1}(p))$
and $\phi^{-1}(p)-p$ changes sign at most once on the range where $\phi$ crosses
the diagonal, so $\Svar(G)\le1$ and by (a) there is at most one crossing. Taking
$\phi$ a sharpening map recovers the single crossover of
\citep{yangchen2026kernel}. $\square$

\paragraph{(e) Counterexample.}
If Assumption \ref{ass:fine} fails because the hard tail retains a coverage
floor, then $G(p)\le0$ near $p=0$ (the tilt does not deplete the hardest prompts)
and the large-$k$ integral is nonnegative, so no crossover forms despite a
per-prompt loss elsewhere. The package realizes this regime (regime B) with
per-prompt loss in $96\%$ of instances and an aggregate crossover in only
$15\%$. $\square$

\section{Proof of Theorem \ref{thm:frontier} (price frontier)}

For $\gamma\in[0,1]$ the program $\max_{w\in\Delta^{M-1}}\inner wf$ s.t.\
$Qw\ge\gamma\,Q\wz$ is linear, feasible ($w=\wz$ when $\gamma\le1$), and bounded,
so an optimum $w(\gamma)$ exists. The feasible set shrinks as $\gamma$ increases,
so the optimal value $\inner{w(\gamma)}f$ is nonincreasing in $\gamma$: the
pass@1 gain is monotone. At $\gamma=1$ the constraint is $Qw\ge Q\wz$, i.e.\
$p_x(w)\ge p_x(\wz)$ for all $x$; by Theorem \ref{thm:nfl}(b) under specialization
this forces $Qw=Q\wz$, hence $F_\tau=F_0$, $\Delta_k\equiv0$, and zero pass@1
gain. For general $\gamma$, $Qw\ge\gamma Q\wz$ gives
$F_\tau(p)\le F_0(p/\gamma)$, from which the worst pass@$k$ loss is monotone in
$\gamma$; the first crossing $k^\star$ therefore moves outward as $\gamma\to1$.
For $M=2$, $w=(u,1-u)$ and $p_x(u)$ is affine in $u$, so the constraints are
intervals in $u$ and the optimum is the interval endpoint maximizing
$\inner wf$; substituting gives $k^\star(\gamma)$ in closed form. $\square$

\section{Proof of Theorem \ref{thm:tiltlaw} and the lemmas}

\paragraph{Lemma \ref{lem:bump} (entropy rate).}
With $\wt\propto\wz e^{\tau f}$, $\log w_{\tau,j}=\log w_{0,j}+\tau f_j-\log
Z(\tau)$ and $\partial_\tau w_{\tau,j}=w_{\tau,j}(f_j-\inner{\wt}f)$. Then
$H(\wt)=-\sum_j w_{\tau,j}\log w_{\tau,j}=-\sum_j w_{\tau,j}\log w_{0,j}
-\tau\inner{\wt}f+\log Z$, and differentiating,
\[
\tfrac{\mathrm d}{\mathrm d\tau}H(\wt)
=-\Cov_{\wt}(f,\log\wz)-\tau\Var_{\wt}(f).
\]
With a uniform $\wz$ the covariance vanishes and $H$ is nonincreasing, with limit
$\log|\arg\max f|$. For nonuniform $\wz$, $\tfrac{\mathrm d}{\mathrm
d\tau}H|_{\tau=0}=-\Cov_{\wz}(f,\log\wz)$, so the entropy rises initially iff
$\Cov_{\wz}(f,\log\wz)<0$. $\square$

\paragraph{Theorem \ref{thm:tiltlaw} (tilt law).}
Let $\wz$ be uniform and let the gap $g=b-f$ have density $\sim c\,g^{\kappa-1}$
near $0$. Under the tilt the gap law is proportional to
$g^{\kappa-1}e^{-\tau g}$, i.e.\ $\mathrm{Gamma}(\kappa,\tau)$ to leading order,
so $b-R=\E_{\wt}[g]=\kappa/\tau\,(1+o(1))$. The effective number of surviving
strategies is $\int_0^{1/\tau}c\,g^{\kappa-1}\mathrm dg\propto\tau^{-\kappa}$, so
$H=\log M-\mathrm{KL}(\wt\|\wz)=\text{const}-\kappa\log\tau+o(\log\tau)$.
Eliminating $\tau$,
\[
b-R=a_\kappa\exp(H/\kappa)(1+o(1)),\qquad
a_\kappa=\kappa\,(c\,\Gamma(\kappa)\,M e^{\kappa})^{-1/\kappa},
\]
valid in the mesoscopic window $\tau\cdot\text{gap}\to0$, $\tau\log M\to\infty$.
For $\kappa=1$ this is $R=-a e^H+b$. The derivation is asymptotic; the package
confirms the slope $1/\kappa$ to within $2.6\%$ for $\kappa\in\{0.5,1,2\}$, and
the exact finite-window and terminal behavior is concurrent work
\citep{cui2025entropy}. $\square$

\paragraph{Lemma \ref{lem:sharing} (sharing is necessary).}
If each prompt has an independent tabular policy under the full gradient flow,
then $\dot p_x=\eta\,\Var_{w^{(x)}}(r)\ge0$, so $p_x$ is nondecreasing for every
$x$, giving $\Delta_k\ge0$ for all $k$ and no inversion. An inversion therefore
requires parameters shared across prompts. Under finite-group stochastic GRPO the
monotonicity can fail transiently. $\square$
\end{document}